\documentclass[conference]{IEEEtran}
\IEEEoverridecommandlockouts
\usepackage{cite}
\usepackage{amsmath,amssymb,amsfonts}
\usepackage{algorithmic}
\usepackage{graphicx}
\usepackage{textcomp}
\usepackage{xcolor}
\usepackage{multirow}
\usepackage{array}
\usepackage{color, soul}

\def\BibTeX{{\rm B\kern-.05em{\sc i\kern-.025em b}\kern-.08em
    T\kern-.1667em\lower.7ex\hbox{E}\kern-.125emX}}
\begin{document}
\title{True or False: Does the Deep Learning Model Learn to Detect Rumors?}

\author{\IEEEauthorblockN{Shiwen Ni, Jiawen Li, and Hung-Yu Kao$^\ast$\thanks{$^\ast$Corresponding author}}
	\IEEEauthorblockA{\textit{Department of Computer Science and Information Engineering} \\
		\textit{National Cheng Kung University}\\
		Tainan, Taiwan\\
		\{P78083033, P78073012\}@gs.ncku.edu.tw, hykao@mail.ncku.edu.tw}

}

\maketitle

\begin{abstract}
It is difficult for humans to distinguish the true and false of rumors, but current deep learning models can surpass humans and achieve excellent accuracy on many rumor datasets. In this paper, we investigate whether deep learning models that seem to perform well actually learn to detect rumors. We evaluate models on their generalization ability to out-of-domain examples by fine-tuning BERT-based models on five real-world datasets and evaluating against all test sets. The experimental results indicate that the generalization ability of the models on other unseen datasets are unsatisfactory, even common-sense rumors cannot be detected. Moreover, we found through experiments that models take shortcuts and learn absurd knowledge when the rumor datasets have serious data pitfalls. This means that simple modifications to the rumor text based on specific rules will lead to inconsistent model predictions. To more realistically evaluate rumor detection models, we proposed a new evaluation method called paired test (PairT), which requires models to correctly predict a pair of test samples at the same time. Furthermore, we make recommendations on how to better create rumor dataset and evaluate rumor detection model at the end of this paper. 
\end{abstract}

\begin{IEEEkeywords}
deep learning, rumor detection, generalization, evaluation method
\end{IEEEkeywords}

\section{Introduction}
Rumors may mislead people's judgments, affect people's lives, and cause adverse effects on society. For example, a mass of rumors about COVID-19 propagating on the Internet while the COVID-19 is spreading globally. And there have been rumors that \emph{Drinking alcohol can kill the new coronavirus in the body}\footnote{https://time.com/5828047/methanol-poisoning-iran/}, and it did cause many people to drink plenty of alcohol and even be hospitalized. Therefore, it is necessary to detect rumors at an early stage. With the development of deep learning, many researchers have tried to train deep learning models for social media rumor detection. The mainstream of deep learning is based on neural networks, and there are also some non-neural based deep learning models \cite{b0}. In this paper, deep learning refers specifically to deep neural networks.

Deep learning has been successfully applied to many natural language processing (NLP) tasks. General rumor detection task is essentially a text binary classifification task \cite{b1}. With the rapid development of NLP technology, current models have achieved surprising performance on many rumor datasets. Under normal circumstances, it is difficult for humans to judge the authenticity of rumors. The current deep learning models can easily reach nearly 90\% accuracy on a rumor dataset \cite{b2,b3,b4,b5,b6,b7}. Does this mean that the models have learned the real ability to detect rumors from the rumor dataset? Obviously this question needs further research and confirmation. And high accuracy on a specific test set does not mean that the model has really learned to detect rumors. 

In this work, we ask the questions: Do models learn to detect rumors? And we subdivide this question into four sub-questions to study separately. 
(1) Does performance on individual rumor datasets generalize to new datasets? (2) Can models detect common-sense rumors? (3) Are the predictions of the model credible and consistent? (4) What does model learn from rumor datasets? 

To answer these questions, first we evaluate models on their generalization ability to out-of-domain examples by fine-tuning BERT-based models on five real-world datasets and evaluating against all five test sets. The experimental results indicate that model performance cannot be well generalized to other unseen datasets. Second, we create a dataset of common-sense rumors to test the trained models and find that the models could not effectively detect common-sense rumors. Third, we analyze some specific cases, and the results show that the predictions of the models are inconsistent. Therefore, the predictions of models may be correct but not credible. Finally, we find that there are serious data pitfalls in Twitter15 and Twitter16 datasets. Those pitfalls lead models to learn absurd knowledge and rules. Based on the experiments and research in this work, we make a certain number of recommendations on how to better create rumor datasets and evaluate rumor detection models at the end of this paper.
\section{Related Works}
Existing rumor detection works are mainly foucus on improving the performance of model on the test set. In contrast, our work focuses on the behavior of models and the real capabilities that models learn from relevant datasets. And our work is inspired by related research and analysis on other NLP task models. \cite{b8} showed through experiments that BERT achieved high performance on the Argument Reasoning Comprehension Task is entirely accounted for by exploitation of spurious statistical cues in the dataset. \cite{b9} explored five QA datasets with six tasks and indicated that models did not learn to generalize well, remained suspiciously robust to wrong data, and failed to handle variations in questions. \cite{b10} found that these QA deep learning models often ignored important question terms. \cite{b11} analyzed the Behavior of Visual Question Answering Models. There are existing perturbation methods meant to evaluate specific behavioral capabilities of NLP models such as logical consistency \cite{b12} and robustness to noise \cite{b13}, name changes \cite{b14}, or adversaries \cite{b15}. Based on behavioral testing in software engineering, \cite{b16} proposed an NLP model testing tool, CheckList, which included a matrix of general linguistic capabilities and test types.

\section{Datasets and Model}
\textbf{Datasets:} We used five rumors or fake news data sets in our experiments: Twitter15, Twitter16, PHEME, GossipCop, PolitiFact. The size and label distribution of the datasets are shown in Table~\ref{t1}. In order to cross-test the data sets, we use only the original rumor text (original Twitter text and news headline) for all datasets, and only take the two labels of "true" and "false". Below we describe each dataset in our experiments:

\begin{table}[t]
	\centering
	\renewcommand\arraystretch{1.15}
		\caption{Overview of the datasets used in this paper.}
	\label{t1}
	\setlength{\tabcolsep}{3mm}{
		\begin{tabular}{|l|r|r|r|c|}
			\hline
			Datasets & \# True & \# False & \# Total & Label: false \% \\ \hline
			Twitter15 & 372 & 370 & 742 & 49.87\% \\ \hline
			Twitter16 & 205 & 205 & 410 & 50.00\% \\ \hline
			PHEME & 3,830 & 1,972 & 5,802 & 33.98\% \\ \hline
			GossipCop & 16,817 & 5,323 & 22,140 & 24.04\% \\ \hline
			PolitiFact & 624 & 432 & 1,056 & 40.90\% \\ \hline
	\end{tabular}}
\end{table}

Twitter15 and Twitter16: Two well-known datasets compiled by \cite{b17}. Each dataset contains a collection of source tweets, along with their corresponding sequences of retweet users. We choose only source tweet text, and "true" and "false" labels as the ground truth. What we need to know is that the ultimate goal of rumor detection is to detect in the early stages of rumors spreading effectively. And rumors are usually only a short paragraph of text in the initial stage (no user comments, no reposting information). Therefore, our work is carried out in the scenario mentioned above.

PHEME\cite{b19}:
This dataset contains a collection of Twitter rumors and non-rumors posted during breaking news. The five breaking news provided with the dataset are as follows:
(1) Charlie Hebdo
(2) Ferguson
(3) Germanwings Crash
(4) Ottawa Shooting
(5) Sydney Siege.

GossipCop and PolitiFact:
The data sets GossipCop and PolitiFact obtained fake news and real news from fact-checking websites \emph{GossipCop.com} and \emph{PolitiFact.com} respectively \cite{b20}. In order to ensure that the text length is similar to other data sets, we only use news headlines as training and test data in the experiment.

\begin{figure}[t]
	\centering
	\includegraphics[width=0.90\linewidth]{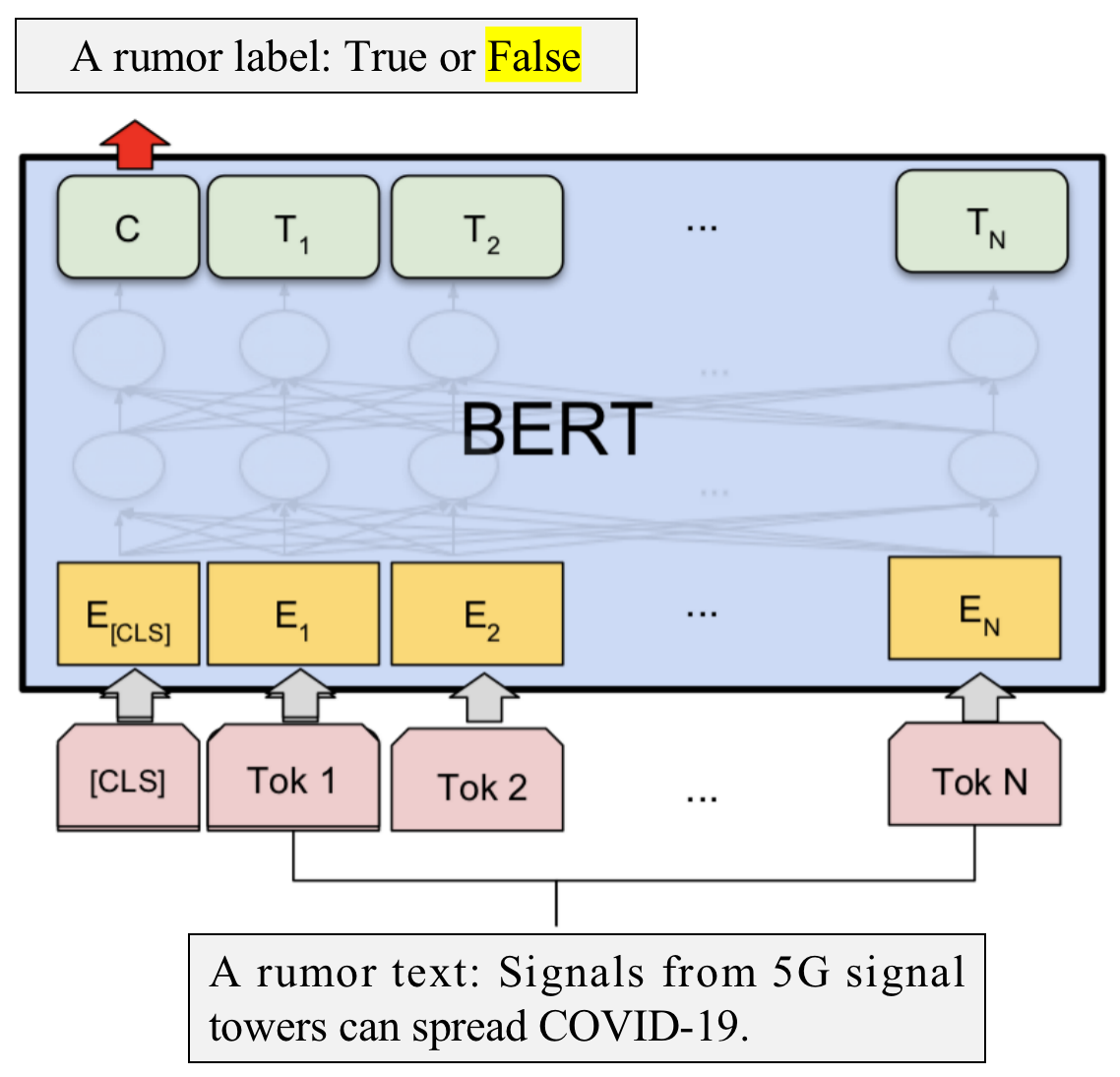}
	\caption{ Detecting a rumor with BERT. Feed a rumor text to the model to learn the representation vector. The final [CLS] vector is then
passed to a linear layer to predicte the label: True or False. }
	\label{f1}
\end{figure}

\textbf{Model}: Note that the purpose of this paper is not to achieve high accuracy on these datasets. In order to facilitate experiments and unify standards, we have chosen a typical deep learning model BERT \cite{b21}, which takes advantage of the self-attention mechanism, and pre-training on a large-scale corpus to learn the general language knowledge and to present state-of-the-art results in many NLP tasks. In this work, all models are initialized from a pre-trained BERT-Base uncased model with 110M parameters. Moreover, each dataset are divided into 70\% training set and 30\% test set. The hyperparameters for fine-tune the models are: Batch Size = 64; Learning Rate =1e-5; Epochs = 8; Max Seq Length =50; Hidden size = 768. The process of using BERT to predict rumors is shown in Fig. \ref{f1}. Feed the model a rumor text, and the model outputs a true or false label.
\begin{table}[]
	\centering
	\label{t2}
	\caption{F1-score of each fine-tuned model evaluated on each test set. The model-on-self baseline is highlighted in \textbf{bold}.}
	\renewcommand\arraystretch{1.2}
	\setlength{\tabcolsep}{0.9mm}{
		\begin{tabular}{|l|c|c|c|c|c|c|}
			\hline
			\multicolumn{2}{|c|}{\multirow{2}{*}{}} & \multicolumn{5}{c|}{Evaluated on} \\ \cline{3-7} 
			\multicolumn{2}{|c|}{} & Twitter15 & Twitter16 & PHEME & GossipCop & PolitiFact \\ \hline 
			\multirow{5}{*}{\begin{tabular}[c]{@{}c@{}}Trained\\ on\end{tabular}} & Twitter15 & \textbf{89.56}& 60.59&38.53&50.34&43.44 \\ \cline{2-7} 
			& Twitter16 & 47.74&\textbf{91.13}&40.54&24.19&28.85 \\\cline{2-7} 
			& PHEME & 45.65&25.81&\textbf{84.34}&50.85&56.32 \\\cline{2-7} 
			& GossipCop & 43.43&47.06&44.29&\textbf{81.57}&57.86 \\\cline{2-7} 
			& PolitiFact & 37.31&32.67&47.48&37.46&\textbf{86.41}\\ \hline
		\end{tabular}
	}
\end{table}
\section{Do Models Learn to Detect Rumors?}
\subsection{Does Performance on Individual Rumor Datasets Generalize to New Datasets?}
For our first experiment, we evaluate the generalizability of models to out-of-domain examples. At present, the vast majority of deep learning works are only tested on a single test set, which shows only a little insight of model's generalization ability. However, generalization ability is important to a model when it is applied in real-world task.

We test generalizability by fine-tuning BERT-based models on each dataset and evaluating against all five test sets. Datasets such as PHEME and GossipCop are unbalanced. Therefore, we choose the high information revealing metric F1-score to measure the performance of the models, and the results are reported in Table II. The rows show a single model’s performance across all five datasets, and the columns show the performance of all the models on a single test set. The model-on-self baseline is indicated in bold.

All models take a great drop in performance when evaluated on an out-of-domain test set. The performance of the models tested on out-of-domain data is even worse than random prediction. This shows that a model's rumor detection performance on an individual dataset does not generalize across datasets. However, the generalization ability of each model is different. The model trained on Twitter15 has an F1-score of 60.59\% when tested on Twitter16.  Similarly, the model trained on Twitter16 achieves 47.74\% of the F1 tested on Twitter15, which is better than the models trained on other datasets. One posssible reason is that the fields and topics of the rumors in Twitter15 and Twitter16 datasets are relatively similar. The models trained on GossipCop is the best generalization ability among the five models. Note that GossipCop and PHEME are the first and second largest in the five datasets, and the generalization ability of the model trained on these two data sets is also the first and second. As can be seen from the above experimental results that more training data can improve the generalization ability of a model.

\begin{table*}[h]
	\centering
		\renewcommand\arraystretch{1.1}
		\caption{Examples of common-sense rumors dataset. There are a total of 200 samples, with 100 rumors and non-rumors each, and the content corresponds to each other.}
		\label{t3}
	\setlength{\tabcolsep}{2.5mm}{
		\begin{tabular}{|c|l|l|}		\hline 
			\# num& Rumors (\# 100) & Non-rumors (\# 100) \\ \hline 
			1&The pet dog next door gave birth to a cat! & Dogs can only give birth to dogs, cats can only  give birth to cats. \\ \hline
			2&Human blood is generally blue. & Human blood is generally red. \\ \hline
			3&In nature, penguins live in the Africa. & In nature, penguins live in Antarctica. \\ \hline
			4&Signals from 5G signal towers can spread COVID-19. & The 5G signal tower does not spread COVID-19. \\ \hline
	\end{tabular}}
	
\end{table*}
\subsection{Can Models Detect Common-Sense Rumors?}
From the experimental results in the previous section, it is proved that the models cannot generalize the high performance well to out-of-domain data. We are wondering if the reason is that the rumors in these datasets are difficult to detect, making it difficult for the model to learn this ability. Therefore, we create a common-sense rumors dataset to verify models’ simple rumor detection ability. The rumors in this dataset are easy for humans to distinguish true and false. We manually collecte and make more than 500 common-sense rumors and corresponding non-rumors by two Ph.D. students and two master students, and then invited 10 people aged 16-40 people (5 males and 5 females) to classify them. We delete the data with more than two people's judgment errors, and finally, we kept 200 pieces of data in our common-sense rumors dataset. Examples of common-sense rumors are shown in Table~\ref{t3}.

The common-sense rumors are adopted to evaluate whether the above models have the ability to detect common-sense rumors. The performance of those models on the original rumor test set and the common-sense rumor test set is shown in Table~\ref{t4}. It can be observed that the accuracy of all models on common-sense rumors is about 50\%, which is basically equivalent to guessing. We further check the the \emph{precision}, \emph{recall} and \emph{f1-score}  of the models, and found that although the accuracy of the models are the same in the common-scense dataset, the other metrics of each model are different. The model fine-tuned on Twitter16 has 98\% of recall in the common-sense rumor test. However, the \emph{precision}, \emph{recall} and \emph{f1-score} of the models fine-tuned on PHEME and GossipCop are all very low. This result shows that the model fine-tuned on Twitter16 is more inclined to predict that the samples are false. The model fine-tuned on PHEME and GossipCop is more inclined to predict that the samples are true. 

The models in Table~\ref{t6} performed very well on the original test set, with an average accuracy of about 87\%. But is the ability of these models to detect rumors really good? The above experimental results prove that these models not only cannot generalize high performance to other rumor test sets, but also basically do not have the ability to detect common-sense rumors.

\begin{table}[h]
	\centering
		\renewcommand\arraystretch{1.1}
			\caption{The performance of the models on the original rumor test set and the common-sense rumor test set. The three criteria of Precision, Recall and F1-score are under the label-False.}
		\label{t4}
	\setlength{\tabcolsep}{1mm}{
	\begin{tabular}{|c|cccc|cccc|}
		\hline
		\multirow{2}{*}{\begin{tabular}[c]{@{}c@{}}Bert\\ fine-tuned on\end{tabular}} & \multicolumn{4}{c|}{\begin{tabular}[c]{@{}c@{}}Test on\\ original rumors\end{tabular}} & \multicolumn{4}{c|}{\begin{tabular}[c]{@{}c@{}}Test on\\ common-sense rumors\end{tabular}} \\ \cline{2-9} 
		& Acc & Pre & Recall & F1 & Acc & Pre & Recall & F1 \\ \hline
		Twitter15 & 89.69& 88.89 & 88.00 & 88.44 & 48.00 & 48.63 & 71.00 & 57.72 \\  \hline
		Twitter16 & 91.13 & 91.94 & 90.48 & 91.20 & 49.00 & 49.49 & 98.00 & 65.77 \\ \hline
		PHEME & 85.22 & 79.77 & 77.63 & 78.68 & 49.50 & 21.23 & 15.32 & 08.66 \\ \hline
		GossipCop &87.87 & 78.51 & 68.89 & 73.36 & 48.50& 18.34 & 10.12 & 06.52 \\ \hline
		PolitiFact & 87.03& 84.55 & 82.54 & 83.53 & 52.00 & 51.30 & 79.00 & 62.20 \\ \hline
	\end{tabular}}

\end{table}

\subsection{Are the Predictions of Models Credible and Consistent?}
We feed a rumor text to a model, and the model will output a label “true” or “false”. The prediction results of a model may be correct, but the results are not necessarily credible because almost all deep learning models are black boxes. In order to know more clearly the ability of the model to detect rumors, the prediction results for each sample are shown in Table~\ref{t5}. We can see that the prediction results of many models are inconsistent. For example, the model PH and GC predict that “\emph{Dogs can only give birth to dogs, cats can only give birth to cats.}” is true, but they also believe that “\emph{The pet dog next door gave birth to a cat!}” is true. In addition, models T15 and T16 predict that the label of “\emph{Human blood is generally blue.}” is false, and it seems that the models are correct. However, the models predict that the label of “\emph{Human blood is generally red.}” is also false. Obviously, the labels of some sample pairs cannot be the same. These examples clearly show that the prediction results of models are not necessarily credible, models just output the label “true" or “false" according to the learned wrong rules. Such a model is like a “mouth” without a “brain”, only output without thinking.
\begin{figure}[t]
	\centering
	\includegraphics[width=0.92\linewidth]{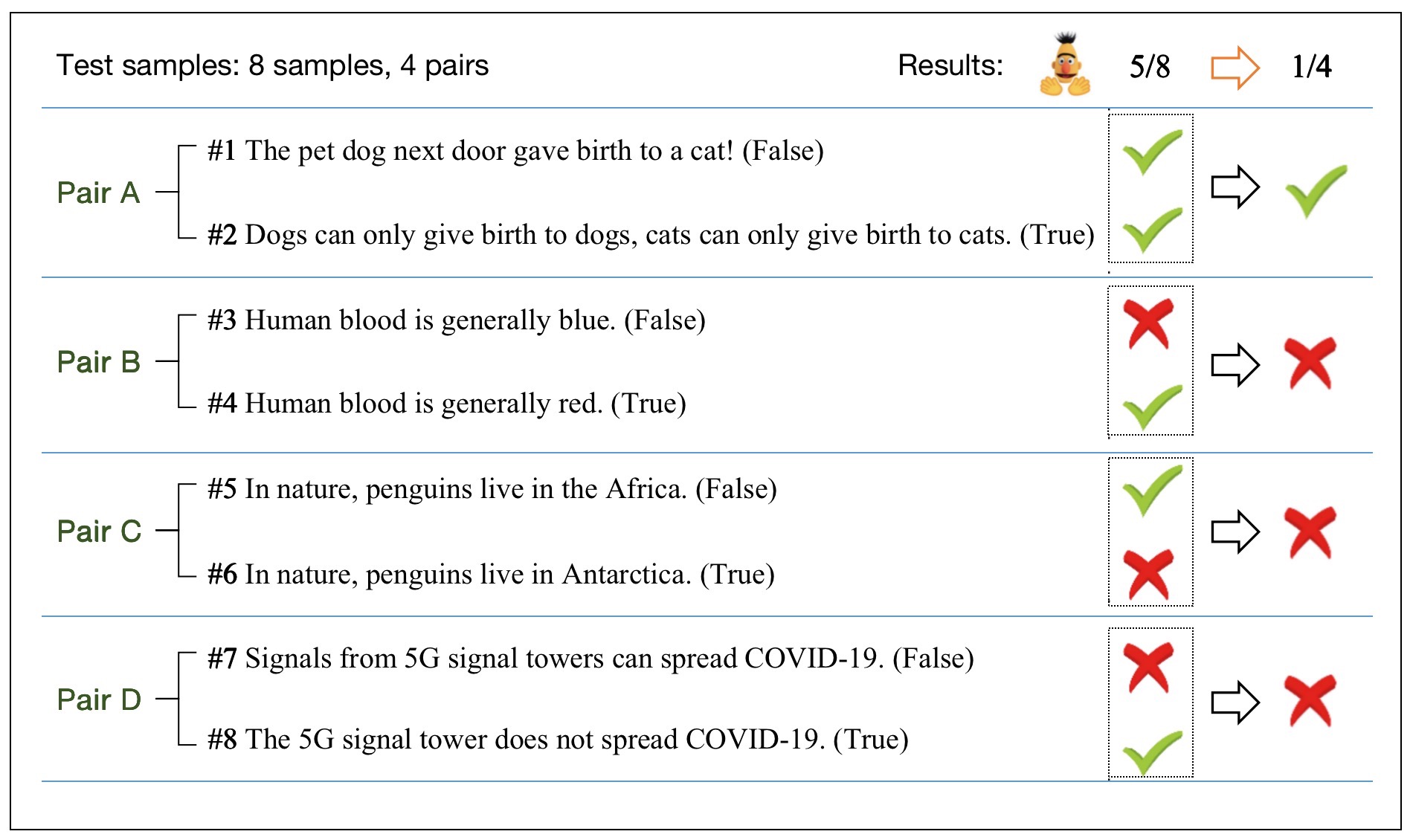}
	\caption{Comparison of standard evaluation method and the proposed PairT evaluation method. Standard evaluation accuracy = 5/8 = 62.5\%; PairT evaluation accuracy = 1/4 = 25\%.}
	\label{f2}
\end{figure}

\begin{table*}[]
	\centering
		\renewcommand\arraystretch{1.1}
		\caption{Models' prediction results case analysis. The models trained on the 5 data sets of Twitter15, Twitter16, Pheme17, GossipCop, and PolitiFact are referred to as T15, T16, PH, GC, and PF respectively. And
			he test samples come from the common-sense rumors dataset. The correct prediction results are \underline{underlined}.}
	\setlength{\tabcolsep}{2mm}{
		\begin{tabular}{|l|c|c|c|c|c|}
			\hline
			\multicolumn{1}{|c|}{} & \multicolumn{5}{c|}{Prediction of models} \\ \cline{2-6} 
			\multicolumn{1}{|c|}{\multirow{-2}{*}{Test samples}} & T15 & T16 & PH & GC & PF \\ \hline 
			The pet dog next door gave birth to a cat! (False) & True & \underline{False} & True & True & \underline{False}  \\
			Dogs can only give birth to dogs, cats can only give birth to cats. (True) & False & False & \underline{True} & \underline{True} & False \\ \hline
			Human blood is generally blue. (False) & \underline{False}  & \underline{False}  & True & True & True \\
			Human blood is generally red. (True) & False & False & \underline{True} & \underline{True} & \underline{True} \\ \hline
			In nature, penguins live in the Africa. (False) & \underline{False}  & \underline{False}  & True & True & \underline{False}  \\
			In nature, penguins live in Antarctica. (True) & \underline{True} & False & \underline{True} & \underline{True} & False \\ \hline
			Signals from 5G signal towers can spread COVID-19. (False) & True & \underline{False}  & True & True & \underline{False}  \\
			The 5G signal tower does not spread COVID-19. (True) & \underline{True} & False & \underline{True} & \underline{True} & False \\ \hline
	\end{tabular}}
	\label{t5}
\end{table*}

Model misjudgment may cause unintended consequences, we need to challenge the rumor detection model higher. The results in Table~\ref{t5}, indicate that the model predicts a sentence as a rumor that does not mean that the model really understands the meaning of the sentence. For example, the model thinks that the sentence "\emph{Dogs can only give birth to dogs, cats can only give birth to cats.}" is true, but model does not understand and learn the true meaning of this sentence. Therefore, the model mistakenly believed that the rumor "\emph{The pet dog next door gave birth to a cat!}" is also true. 

In order to more realistically evaluate the performance of the rumor detection model, we propose a new evaluation method called paired test (\textbf{PairT}) in this paper. In this new evaluation method, models should test samples in pairs such as [A \& A$’$], the A is the original rumor text, and the A$’$ is a new text created manually. The important point is that the hidden knowledge and label contained in sample A are opposed to sample A$’$. And the model needs to predict samples A and A$’$ correctly at the same time. If a model predicts that the labels of A and A$’$ are the same, it means that this model is unreliable, because the prediction results are inconsistent. As shown in Fig.~\ref{f2}, there are eight test samples, that is, four pairs of A = [\#1 \& \#2], B = [\#3 \& \#4], C = [\#5 \& \#6], D = [\#7 \& \#8]. Assuming that a model predicts the five samples \#1, \#2, \#4, \#5 and \#8 correctly, and the three samples of \#3, \#6 and \#7 wrong, the accuracy rate calculated according to the standard evaluation method is 62.5\%. But according to our new evaluation method \textbf{PairT}, only one pair, the Pair A = [\#1 \& \#2], has been correctly predicted. So the accuracy of the
model is 25\%. We hope that the output of model will not contradict each other, and make inferences based on the true knowledge learned as much as possible, rather than ridiculous shortcut rules. This will pose a higher challenge to models so that models can be better applied in the real-world.

\begin{table}[]
	\centering
		\renewcommand\arraystretch{1.1}
		\caption{By calculating the s and b of each word, we found out the words “Obama”, “Paul” and “Sydney” that satisfy $s\geq s_{min} = 0.8,  b\geq b_{min} = 5\%$.}
	\label{t6}
	\setlength{\tabcolsep}{2mm}{
	\begin{tabular}{|c|c|c|c|c|}
		\hline
		\multirow{2}{*}{} & \multicolumn{2}{c|}{Twitter15} & \multicolumn{2}{c|}{Twitter16} \\ \cline{2-5} 
		& Obama & Paul & Obama & Sydney \\ \hline
		 Strength $s$  & 0.95 & 0.92 & 0.96 & 0.89 \\ \hline
		Breadth $b$ & 5.4\% & 11.2\% & 5.0\% & 11.7\% \\ \hline
	\end{tabular}}
\end{table}

\sethlcolor{yellow} 
\begin{table*}[]
	\centering
			\renewcommand\arraystretch{1.03}
		\caption{Original and Adversarial data. The label of the adversarial data relative to the original data must be changed. The place where the text has changed is \hl{highlighted}.}
	\label{t7}
	\setlength{\tabcolsep}{2mm}{
	\begin{tabular}{|m{6cm}|l|m{6cm}|l|}
		\hline
		\multicolumn{2}{|c|}{Original} & \multicolumn{2}{c|}{Adversarial} \\ \hline
		\multicolumn{1}{|c|}{Text} & label & \multicolumn{1}{c|}{Text} & label \\ \hline 
		Obama \hl{says}, “America doesn't want any stay-at-home-moms! Enough is enough!” & False & Obama \hl{does not say}, “America doesn't want any stay-at-home-moms! Enough is enough!” & True \\ \hline
		r.i.p to the driver that died with Paul that \hl{no one cares} about because he \hl{wasn't} famous. & True & r.i.p to the driver that died with Paul that \hl{many poeple care} about because he \hl{was} famous. & False \\ \hline
		Hostage situation erupts in Sydney cafe, Australian prime minister says it \hl{may be} “politically motivated” & True & Hostage situation erupts in Sydney cafe, Australian prime minister says it \hl{could not be} “politically motivated” & False \\ \hline
	\end{tabular}}
\end{table*}

\begin{table}[h]
	
		\renewcommand\arraystretch{1.1}
			\caption{Results for BERT on the adversarial test. O, P and S stand for “Obama”, “Paul” and “Sydney” respectively. The dataset Twitter15 contains the words “Obama” and “Paul”; the dataset Twitter16 contains the words “Obama” and “Sydney”. Declining accuracy are highlighted in \textbf{bold}.}
		\label{t8}
			\centering
\setlength{\tabcolsep}{2mm}{
		\begin{tabular}{|m{3.5cm}|c|c|}
			\hline
			& Twitter15 & Twitter16 \\ \hline 
			Original test set & 89.69\% & 91.13\% \\ \hline
			Adversarial (O) & 84.75\% & 81.45\% \\ \hline
			Adversarial (O \& P or S) & 73.09\% & 68.55\% \\ \hline
			Declining accuracy & \textbf{16.60\%} & \textbf{22.58\%} \\ \hline
	\end{tabular}}
\end{table}

\subsection{What Does Model Learn from Rumor Datasets?}
The above experimental results indicate that the models do not learn to detect rumors well. In order to know what the model has learned, we use the word-level attention mechanism to analyze the words that models mainly focus and breadth $b$ of its own signal, where the strength $s$ is the average attention weight of the word, and the breadth $b$ is the proportion of the data containing the word to all the data. In order to find the clue words that have a serious impact on the model, we set the thresholds of $s$ and $b$ to $s_{min} = 0.8, b_{min} = 5\%$. Finally, three clue words were found on Twitter15 and Twitter16, as shown in Table~\ref{t6}. A single word occupies about 90\% of the attention weight in a sentence, which is obviously unreasonable. After checking the original datasets, we found that the reason is that the distribution of these words is very uneven. For example, the word “Obama” almost only appears in the samples whose label is “False”. Therefore, the model trained on these two datasets is likely to take shortcuts to achieve high performance. 

For the model trained on Twitter15, it tends to predict “false” when a test sentence contains the word “Obama”; In contrast, the model tends to predict “true” while a test sentence includes the phrase “Paul”. And for the model trained on Twitter16, it also tends to predict “false” when a test sentence contains the word “Obama”; In contrast, the model tends to predict “true” while a test sentence contains the word “Sydney”. This phenomenon is obviously illogical and unreasonable. In order to verify this finding, an adversarial dataset is established. We reverse the meaning of the data containing “Obama”, “Paul” and “Sydney” in the Twitter15 and Twitter16 datasets, and change the label of the data. The adversarial data is shown in Table~\ref{t7}. We test the performance of the models on the adversarial dataset, and the experimental results are shown in Table~\ref{t8}. When only the data containing “Obama” is modified, the accuracy of the model on the two datasets is 84.75\% and 81.45\%, which is about 5 and 10\% lower than the original accuracy, respectively. By further modifying the sentences containing “Paul” and “Sydney”, the accuracy of the model drops to 73.09\% and 68.55\%, respectively, and the accuracy drops by 16.60\% and 22.58\% compared to the original accuracy, respectively. Those accuracy drops are huge and alarming, because the words “Obama”, “Paul” and “Sydney” have great strength $s$ and breadth $b$ at the same time. Just because of the uneven distribution of a word, the results of the model will be severely affected. That is a cautionary tale for us that we need to be more cautious when creating a rumor dataset.

\begin{figure}[]
	\centering
	\includegraphics[width=0.95\linewidth]{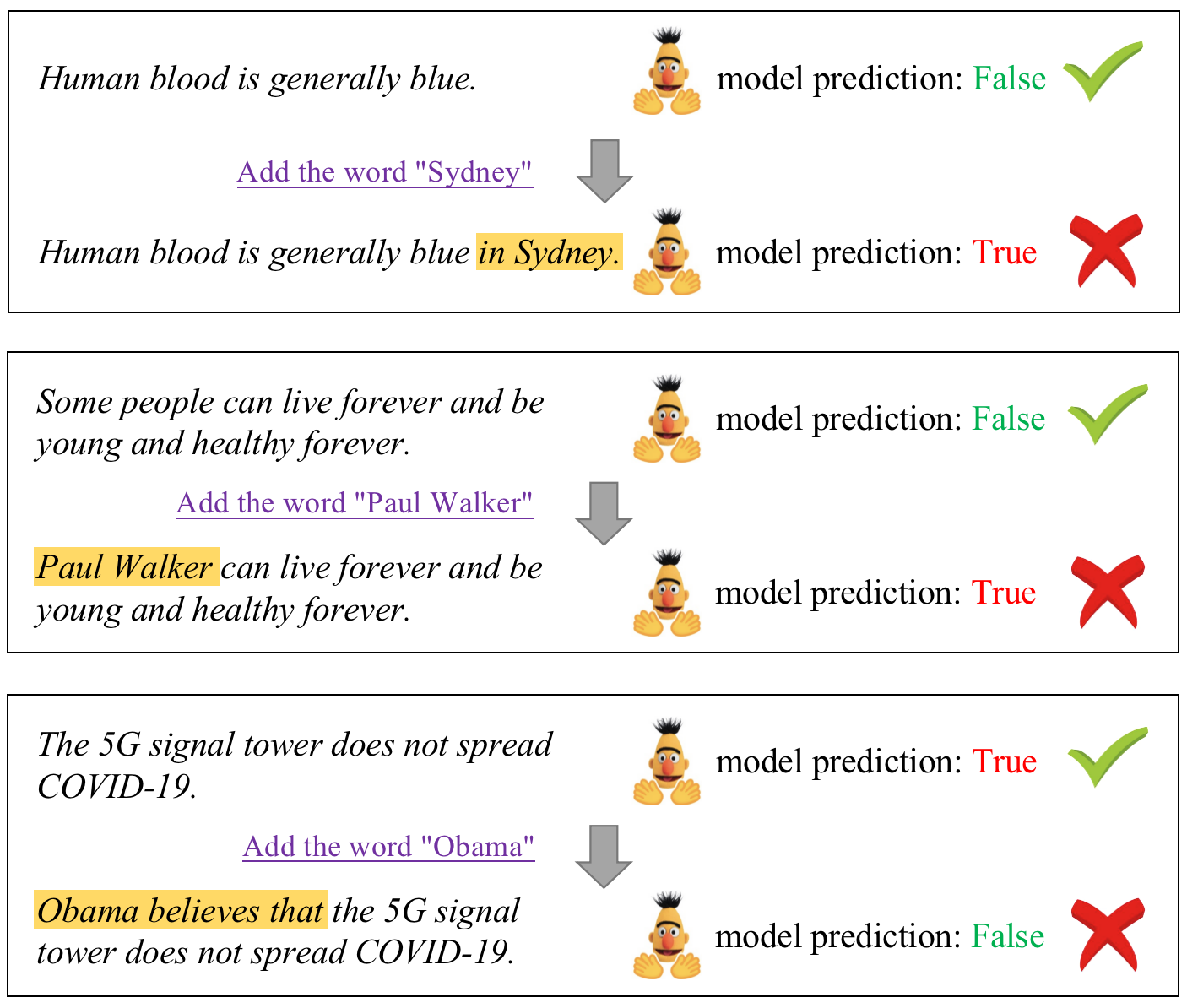}
	\caption{Misleading the model to make wrong predictions through spurious clues. The test model is Bert fine-tuned on dataset Twitter15, and the text modification is highlighted.}
	\label{fig:2}
\end{figure}

In addition, we analyzed three specific cases to point out the absurd knowledge and rules learned by the model. As we can see from Fig.~\ref{fig:2}, the label predicted by the model for the sentence “\emph{Human blood is generally blue.}” is False. If the sentence is changed to “\emph{Human blood is generally blue in Sydney.}”, the label predicted by the model will become True, which is obviously unreasonable. The model will output the label “True” as long as it sees the word “Sydney”. Similarly, we change the sentence “\emph{Some people can live forever and be young and healthy forever.}” to “\emph{Paul Walker can live forever and be young and healthy forever.}”, the model prediction result will also change from “False” to “True”. In short, if the text contains “Sydney” or “Paul”, the model will predict “True” with a high probability. On the contrary, if the text includes “Obama”, the model will output “False” with a high probability. The root cause of the model taking shortcuts or deception is that those two datasets, Twitter15 and Twitter16, left data pitfalls in the collection and creation process. There are two main reasons for serious data pitfalls. First, the range of events when collecting data is small, which leads to high coverage of certain (“Obama”, “Paul”, “Sydney”) words in the data set in all data. Second, the data containing these words (“Obama”, “Paul”, “Sydney”) is seriously unbalanced. For example, in the process of data collection, more than 90\% of the data containing “Obama” are labeled as false. In the data collection process, these issues are not difficult to avoid.

\section{Conclusion}
In this work, we conducted a series of experiments using five public datasets and the pre-training model BERT, and found that the seemingly high-performance models do not really learn to detect rumors. When dealing with a task, deep learning models are actually more likely to take shortcuts and cheat than humans. The good performance of a model may be because it learns some hidden clues and simple rules in the dataset. The shortcomings in datasets and evaluation methods make it difficult to judge whether  models learn to detect rumors or not. Based on our work, we offer the following recommendations to researchers who create rumor datasets or train and evaluate rumor detection models:

\textbf{Do not just focus on the accuracy, precision, recall and f1-score of models.} Training a rumor detection model should focus not only on improving the accuracy, precision, recall and f1-score of the model but also on the behavior of the model, the credibility of model predictions, and the interpretability of model results.

\textbf{Test out-of-domain examples for the generalization ability of models.} A model has better generalization ability, and its application in the real world will be more valuable. A new rumor detection model should evaluate the performance across multiple related datasets.

\textbf{Challenge the models and use stricter evaluation criteria for them.} Evaluating on some easy test sets will exaggerate our judgments about the real capabilities learned by models. Rumor detection models can be evaluated using our proposed new evaluation method PairT.

\textbf{Create datasets carefully to avoid data pitfalls and ensure that models do not take shortcuts and cheat.} When creating a new rumor dataset, prevent a large number of unbalanced data clues, such as “Obama”, “Paul” and “Sydney” found in the paper. Try to eliminate such data pitfalls to avoid model taking shortcuts and cheat.

\section*{Acknowledgment}
This work was funded in part by Qualcomm through a Taiwan University Research Collaboration Project and in part by the Ministry of Science and Technology, Taiwan, under grant MOST 110-2221-E-006-001 and NCKU B109-K027D. We thank to National Center for High-performance Computing (NCHC) for providing computational and storage resources.

\end{document}